# Predicting BWR Criticality with Data-Driven Machine Learning Model

Muhammad Rizki Oktavian*, Anirudh Tunga[†], Jonathan Nistor[†], James Tusar[§], J. Thomas Gruenwald[†], Yunlin Xu*

rizkiokt@purdue.edu

*School of Nuclear Engineering, Purdue University, 516 Northwestern Ave., West Lafayette, IN 47906
[†]Blue Wave AI Labs, 1281 Win Hentschel Blvd, West Lafayette, IN 47906
[§]Nuclear Fuels Department, Constellation, 1310 Point Street, Baltimore, MD 21231

## INTRODUCTION

One of the challenges in operating nuclear power plants is to decide the amount of fuel needed in a cycle. Large-scale nuclear power plants are designed to operate at base load, meaning that they are expected to always operate at full power. Economically, a nuclear power plant should burn enough fuel to maintain criticality until the end of a cycle (EOC) [1]. If the reactor goes subcritical before the end of a cycle, it may result in early coastdown as the fuel in the core is already depleted. On contrary, if the reactor still has significant excess reactivity by the end of a cycle, the remaining fuels will remain unused. In both cases, the plant operator may lose a significant amount of money.

For a Boiling Water Reactor (BWR), it is important to assure the core is adequately subcritical with control rods under cold conditions [1]. Due to the boiling phenomena in BWR, it is not feasible to add soluble boron to the reactor coolant like other light water reactor designs. This condition restricts the excess reactivity of BWR at the beginning of a cycle (BOC). Practically, BWR has around 0.8% $\Delta k$ of excess reactivity at BOC to ensure this requirement is fulfilled [2].

A standard method to estimate excess reactivity in reactor operation is to utilize a core simulator to run calculation based on reactor operational conditions. For General Electric BWRs, a core simulator is used to estimate this parameter [3]. However, it is possible to obtain inaccurate predictions for some cases due to the nature of approximation in physics-based numerical methods.

This work proposes an innovative method based on a data-driven machine learning model to estimate the excess criticality of a boiling water reactor. Similar works related to the use of the data-driven machine learning method in predicting parameters in BWRs have been conducted in [4] and [5]. Both works predict the amount of moisture carryover (MCO) from the reactor with excellent accuracy.

To predict the $k$ of the reactor, this research employs the data based on the Local Power Range Monitoring (LPRM) system output. The LPRM is an in-core detector used to measure local neutron flux at various radial and axial locations in a reactor [6]. Since the local flux is influenced mostly by control rod patterns and other operational parameters, it is important to consider including those parameters in the machine learning model as well. All those parameters are also recorded along with the output from the LPRM data.

It is true that the core eigenvalue is directly caused by the change in the uranium and gadolinia concentration in the reactor and the increasing exposure. We have built a machine learning model based on those parameters in our previous project. However, it is of interest in this work to use LPRM measurements to find the possible correlation with the core parameter. These measurements can be recorded online during reactor operation, thus accurate predicting capability is preferred for this data.

## METHODOLOGY

### Dataset

The dataset used in this work is provided by a large Nuclear Power Provider on one of their BWR nuclear power stations. The dataset consists of operational data from the reactor, including the LPRM measurement readings. The criticality data is obtained daily from the core simulator using the recorded reactor operational conditions to estimate the $k$ on All Rod Out (ARO) positions. The progression of this value along the reactor operation days is presented in Fig. 1 for cycle 20 to cycle 23.

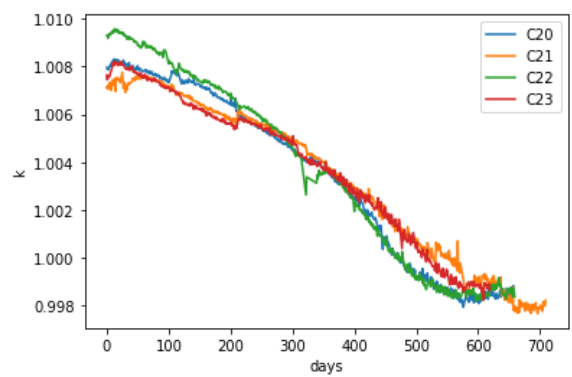

Fig. 1. Progression of $k$ in various reactor cycles.

The number of available data for each cycle is different with the newer cycle having more data than the older cycle. The available datapoints for each cycle are presented in Fig. 2. In that figure, the dataset is dominated by the Cycle 23 data with more than 7,000 datapoints. Other cycles make up the rest of the dataset, making the total datapoints to be around 10,000 data.

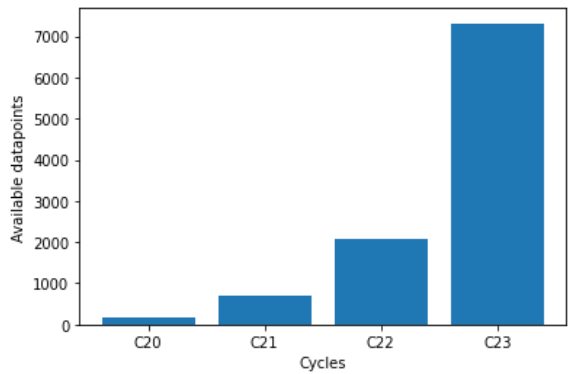

Fig. 2. Available operational data points for cycles 20-23 of the BWR station we analyze.

Based on the available datapoints, we use them as inputs for our machine learning model using all the features that might be useful to predict $k$. The features can be divided based on the data shape or dimensions as zero-dimensional, two-dimensional, and three-dimensional features. This list of features utilized in this work can be found in Table 1.

TABLE I. Features Utilized from the Available Data

| 3D Features | Shape: (25,30,30,4) |
|---|---|
| 0 | Interpolated LPRM Readings |
| 1 | Nodal Iodine Worth |
| 2 | Nodal Power |
| 3 | Nodal Xenon Worth |
| **2D Features** | **Shape: (30,30,1)** |
| 0 | Control Rod Pattern |
| **0D Features** | **Shape: (5)** |
| 0 | Core Dome Pressure |
| 1 | Core Flow |
| 2 | Core Inlet Subcooling |
| 3 | Thermal Power |
| 4 | Core Xenon Worth |

**LPRM Data Interpolation**

The LPRM data are one of the most important measurement data in BWR. For the station we analyze, there are 43 radial locations of LPRM strings, with each string containing four detectors spaced at three-foot intervals. In total there are 172 local flux points measured in the reactor. Altogether, these data represent the neutron thermal flux shape in the reactor [6].

Since the LPRM data are only available in a few local points in the reactor, it is better to have more interpolated data to cover more locations in the reactor geometry. This can be done using either a standard interpolation method, like piecewise constants or 3D linear interpolation, or higher order method. This work implements 3D spline interpolation to find the local points not covered by LPRM measurements. The interpolation results of the neutron flux based on the available LPRM datapoints are presented in Fig. 3.

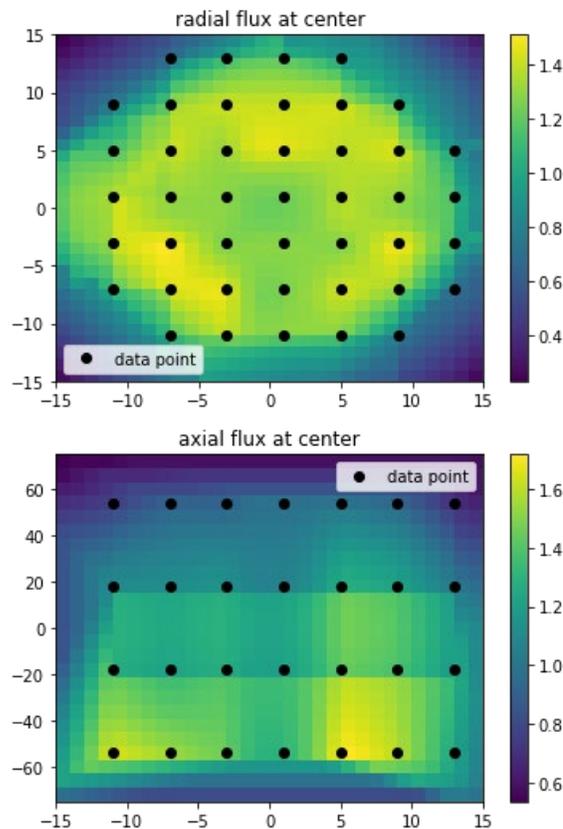

Fig. 3. Example of interpolation results on LPRM data, resulting in higher resolution radial and axial normalized neutron flux distribution.

**Machine Learning Model**

Fig. 4 presents the network architecture of the machine learning used in this work. It consists of three different branches of models based on input dimensionality. The multi-dimensional input data entered the 2D and 3D convolutional neural networks (CNN) before being flattened into one-dimensional arrays and concatenated with other input data. Most of the CNN applications are in the field of image recognition, so the use of CNN in reactor applications are quite innovative. The capability of CNN to extract the spatial dependence of features has an immense potential to handle high dimensional data like nodal flux and rod pattern.

After all the input data are concatenated into a one-dimensional array, it enters two layers of fully connected neural networks with 1024 nodes each. Some dropout layers are also added to reduce the model overfitting. The training, validation and testing process is performed using GPU computing system to accelerate the training process.

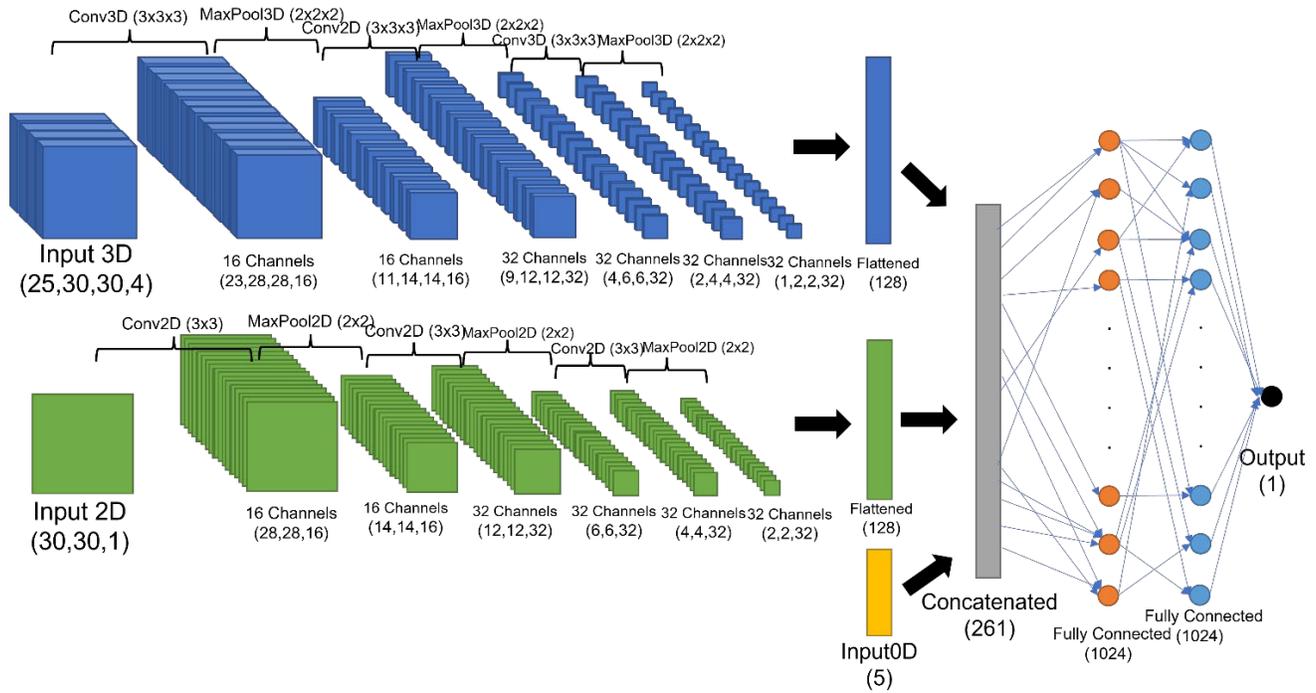

Fig. 4. The network architecture of the machine learning model. The high-dimensional input features are processed using convolutional neural networks, before being flattened and concatenated with other input features. Note that the number in the parentheses is the array shape.

**Training-Validation-Test Split**

Based on the available datasets, training, validation, and test data are obtained with two different split methods:
1. Random Split: Training-Validation-Test is taken randomly with a 70-15-15 ratio.
2. Cycle-based Split: Cycle 20 and 23 data are used for the training set, cycle 21 for the validation set, and cycle 22 for the test set.

Both split methods will be compared to see the performance of the machine learning model to predict $k$. The random split is not realistic in terms of the actual practice. These datasets will be used to judge the learning capability of the model in ideal data conditions, i.e., training, validation, and test data come from the same data distribution. The real test to the model will be in the cycle-based split, which is more applicated to the real-world problems.

## RESULTS

The experiments involved running the models to train, validate and test datasets using two different split methods. The model was evaluated with Mean Squared Error (MSE) as the loss function. It is trained using 128 batches of the data for each step with a maximum of 1000 epochs. Early stopping criteria were set to avoid overfitting the training data. In addition to the MSE, $R^2$ value or the coefficient of determination was also used to determine the network performance on the datasets.

**Random Split Results**

The training and validation loss of the model to the random split datasets can be observed in Fig. 5. The training loss and validation loss reduce along with the epoch until reached a plateau and stopped early to avoid overfitting. The validation loss is comparable with the training loss at the end of the training epoch. This suggests that the model can generalize the datasets very well.

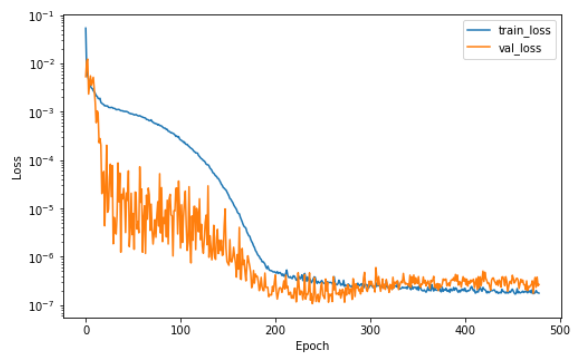

Fig. 5. Training and validation progression along with training epoch for random split datasets.

In Table II, we can see that the model can provide an excellent agreement with all training, validation, and testing datasets with exceedingly small MSE, and the $R^2$ value that are close to 1.0. This means the model can predict the target well. In Fig. 6, the prediction results are mostly close to the target value, except for some data near the end of the cycle.

TABLE II. Accuracy Table of Machine Learning Model with Random Split Datasets

| Case | MSE | $R^2$ value |
|---|---|---|
| Training | $2.611 \times 10^{-7}$ | 0.977 |
| Validation | $2.668 \times 10^{-7}$ | 0.977 |
| Testing | $2.762 \times 10^{-7}$ | 0.976 |

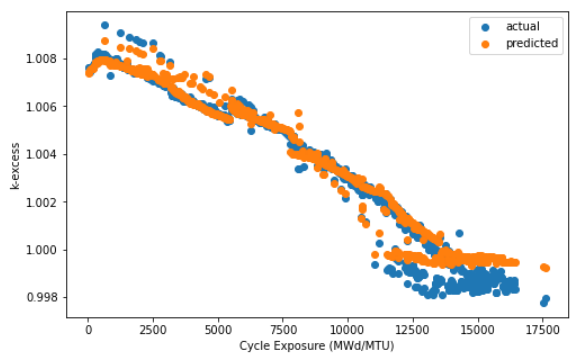

Fig. 6. Predicted results compared to the actual data on the testing dataset using random split.

**Cycle-based Split Results**

The training process in Fig. 7 for cycle-based split datasets suggests that the model cannot generalize enough to predict the validation set accurately. This is further confirmed in Table III as the MSE and $R^2$ value on validation and testing sets are significantly worse than the ones on the training sets.

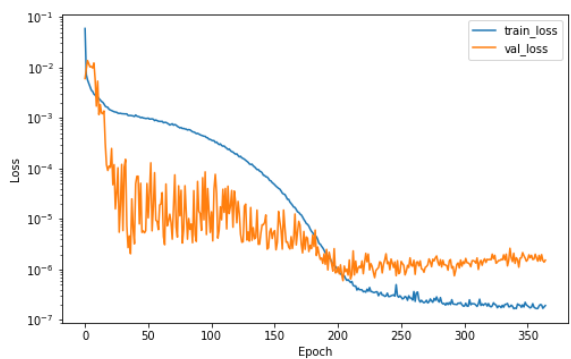

Fig. 7. Training and validation progression along with training epoch for cycle-based split datasets.

TABLE III. Accuracy Table of Machine Learning Model with Cycle-based Split Datasets

| Case | MSE | $R^2$ value |
|---|---|---|
| Training | $5.117 \times 10^{-7}$ | 0.944 |
| Validation | $1.532 \times 10^{-6}$ | 0.850 |
| Testing | $3.519 \times 10^{-6}$ | 0.637 |

Fig. 8 shows more about the model performance on the testing set. The model underpredicts at the beginning of the cycle and then overpredicts the reference data near the end of the cycle. This can be understood since the training data are dominated by Cycle 23 data, the prediction also tends to follow this cycle data.

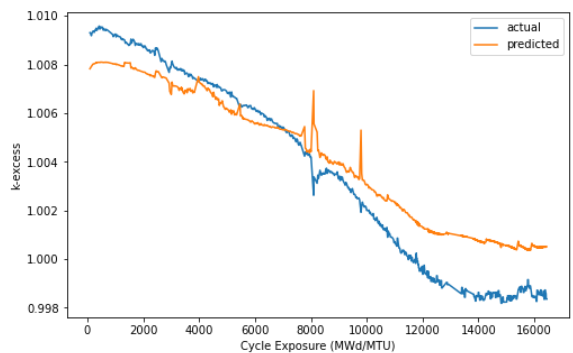

Fig. 8. Predicted results compared to the actual data on the testing dataset using the cycle-based split method.

**CONCLUSION**

The results of this work prove the concept of eigenvalue predictive modeling using deep learning approach. Since this is still preliminary development of this work, there is a room for improvement to handle the problems occurred, especially related to the training data. Future development to the methods might include the addition of physics constraints, data augmentation, and more extensive reactor cycle data.

**ACKNOWLEDGMENTS**

This research was supported by an appointment to the 2022 summer internship program at Blue Wave AI Labs.

**REFERENCES**
1. P. J. TURINSKY, G. T. PARKS, "Advances in Nuclear Fuel Management for Light Water Reactors," *Advances in Nuclear Science and Technology*, **26** (2002).
2. O. OZER, "Optimum Cycle Length and Discharge Burnup for Nuclear Fuel," Technical Report 1003217, EPRI and U.S. Department of Energy (2002).
3. J. E. Wieging, "Validation of TGBLA/PANACEA code package to Commonwealth Edison BWRs," *Proc. of the 1992 Topical Meeting on Reactor Physics*, volume 1 (1992)
4. H. WANG, J. T. GRUENWALD, J. TUSAR, R. VILIM, "Moisture-carryover performance optimization using physics-constrained machine learning," *Progress in Nuclear Energy*, **135,** 103691 (2021).
5. H. WANG, et. al. (2019), "Machine-Learning Analysis of Moisture Carryover in Boiling Water Reactors," *Nuclear Technology,* **205,** 1003-1020 (2019).
6. "General Electric System Technology Manual: Local Power Range Monitoring System," USNRC HRTD (2011).